%% file: rwsl.tex
\documentclass[conference]{IEEEtran}
\IEEEoverridecommandlockouts
\usepackage{bm}
\usepackage{comment}
\usepackage{lipsum}

\usepackage{algorithm}
\usepackage{algpseudocode}

\usepackage{amsthm}

\newtheorem{claim}{Claim}

\usepackage{graphicx,float}
\usepackage{adjustbox}
\usepackage{caption}
\usepackage{subcaption}  
\usepackage{wrapfig}

\usepackage{cite}
\usepackage{amsmath,amssymb,amsfonts}
\usepackage{graphicx}
\usepackage{textcomp}
\usepackage{xcolor}
\def\BibTeX{{\rm B\kern-.05em{\sc i\kern-.025em b}\kern-.08em
    T\kern-.1667em\lower.7ex\hbox{E}\kern-.125emX}}

\usepackage{fancyhdr}
\usepackage{kantlipsum}
\fancypagestyle{plain}{
\fancyhf{}
\fancyhead[C]{Conference on \LaTeX} %% C or L or R.
}
\usepackage{eso-pic}

\begin{document}

\AddToShipoutPictureBG*{
\AtPageUpperLeft{
\setlength\unitlength{1in}
\hspace*{\dimexpr0.5\paperwidth\relax}%% change \dimexpr0.5\paperwidth\relax appropriately
\makebox(0,-0.75)[c]{\textbf{2022 IEEE/ACM International Conference on Advances in Social Networks Analysis and Mining (ASONAM)}}}}

\title{Deep Graph Clustering with Random-walk based Scalable Learning\\
}

% \author{\IEEEauthorblockN{1\textsuperscript{st} Xiang Li}
% \IEEEauthorblockA{\textit{Computer Science and Engineering} \\
% \textit{The Ohio State University}\\
% Columbus, USA \\
% li.3880@osu.edu}
% \and
% \and
% \and
% \and

\author{\IEEEauthorblockN{Xiang Li }
\IEEEauthorblockA{\textit{Computer Science and Engineering} \\
\textit{The Ohio State University}\\
li.3880@osu.edu} \\
\IEEEauthorblockN{Rajiv Ramnath }
\IEEEauthorblockA{\textit{Computer Science and Engineering} \\
\textit{The Ohio State University}\\
	ramnath@cse.ohio-state.edu}
	\and 
	\IEEEauthorblockN{Dong Li }
\IEEEauthorblockA{\textit{Computer Science Department} \\
\textit{Kent State University}\\
dli12@kent.edu} \\
\IEEEauthorblockN{Gagan Agrawal}
\IEEEauthorblockA{\textit{Computer and Cyber Sciences } \\
\textit{Augusta University}\\
	gagrawal@augusta.edu} 
\and 
\IEEEauthorblockN{Ruoming Jin }
\IEEEauthorblockA{\textit{Computer Science Department} \\
\textit{Kent State University}\\
rjin1@kent.edu} \\

}

\maketitle

\IEEEoverridecommandlockouts
\IEEEpubid{\parbox{\columnwidth}{\vspace{8pt}
\makebox[\columnwidth][t]{IEEE/ACM ASONAM 2022, November 10-13, 2022}
\makebox[\columnwidth][t]{978-1-6654-5661-6/22/\$31.00~\copyright\space2022 IEEE} \hfill} \hspace{\columnsep}\makebox[\columnwidth]{}}
\IEEEpubidadjcol

\begin{abstract}
Interactions between (social) entities can be frequently represented by an attributed graph, and node clustering in such graphs has received much attention lately. 
Multiple efforts have  successfully applied Graph Convolutional Networks (GCN), though with some limits on accuracy as  GCNs have been shown to 
suffer from over-smoothing issues. Though other methods (particularly those based on Laplacian Smoothing) have reported better accuracy, a fundamental limitation 
of  all the work is a  lack of scalability.  This paper addresses this open problem by relating  the Laplacian smoothing to the  Generalized PageRank, and applying  a random-walk based algorithm  as a scalable graph filter. This forms the basis for our scalable deep clustering algorithm, RwSL. Using  6 real-world datasets and 6 clustering metrics, we show that RwSL achieved improved results over several recent baselines. Most notably,  by demonstrating execution of RwSL on a graph with 
1.8 billion edges using only a single GPU.
We show that RwSL  can  continue to scale, unlike  other existing deep clustering frameworks.
\end{abstract}


\baselineskip=0.97\normalbaselineskip 

\input{text/intro}

\input{text/related_work}

\input{text/methods}
\input{text/analysis}

\input{text/experiment}

\input{text/Conc}

\subsection*{Acknowledgements:} This work was partially supported by NSF grants 2007793, 
2034850, and 2131509.

\bibliographystyle{unsrt}
\bibliography{rwsl}

\end{document}

%% file: text/intro.tex
\vspace*{-2ex} 
\section{Introduction} 
\label{sec:introduction} 

Many interactions between (social)  entities  are commonly 
represented as {\em attributed graphs}  (graphs with {\em  node attributes}). 
%There has been much interest in 
%developing techniques for tasks like node classification, node clustering, 
%and link prediction on such graphs, driven by applications like 
%   analysis of citation networks \cite{kipf_2017_semi_GCN}, social network \cite{Lazer_2009_life_network}, and 
 %recommendation systems \cite{Ying_2018_GCN_web}.
Node clustering~\cite{Kmeans_1979}, when applied to  an attributed graph,  involves  the challenge of jointly learning from the non-Euclidean graph structure and the high-dimensional node attributes. 
%Self-supervised learning \cite{self_supervised_CL_Jaiswal_2021}, which has the   capability of incorporating self-defined ``pseudo-labels''  can naturally handle unsupervised learning tasks such as clustering.
Deep clustering methods \cite{DONG_2021_DP_survey}, which  
integrate the clustering objective(s) with deep learning (particularly 
 Graph Convolutional Networks (GCNs) \cite{kipf_2017_semi_GCN,review_GNN_zhou_2018}), have been investigated by several researchers. A majority of GCN based frameworks for node clustering are based on Graph Autoencoder (GAE) \cite{kipf2016variational_VGAE}. 
 Especially, advanced clustering performance was  achieved from 
 GCN based 
 %self-supervised 
learning frameworks \cite{velickovic_2019_DGI,Bo_2020_structural_deep}.
 
  However, the introduction of GCN in deep clustering  also brought in the inherent 
limitations  of GCN. Previous work  has demystified GCN as a special case of Laplacian smoothing -- with the implication that 
it will suffer from the over-smoothing issue which can 
limit the learning ability~\cite{Li_Deeper_into_GCN_2018}. 
To overcome GCN drawbacks, other efforts use carefully designed graph smoothing filters to ease node clustering tasks, such 
as those used in Simple Graph Convolution (SGC)~\cite{SGC_2019}, Adaptive Graph Convolution (AGC)~\cite{ijcai2019_AGC}
%method to exploit high-order graph convolutions to capture global cluster structure.
 and  Simple Spectral Graph Convolution  (SSGC)~\cite{ssgc_2021}. 
%from the Markov Diffusion Kernel to aggregate k-step diffusion matrices. 
It turns out, however,  graph smoothing filters based methods still suffer from the same scalability issues (as do the  GCN based approaches) 
since the adjacency matrix of the entire graph is  used for a sparse matrix multiplication, resulting in high computational and memory requirements. 
%which are prohibitive for large-scale graphs. 
 
To address the open problem of developing scalable and accurate node clustering of an attributed graph, 
in this paper we derive  that 
the Laplacian smoothing filter~\cite{Li_Deeper_into_GCN_2018} is  a form of 
Generalized PageRank (GPR) \cite{PanLi_2019seed_expansion}. Further observing that GPR can be effectively approximated by a scalable and parallelizable random-walk based algorithm \cite{cwdlydw2020gbp}, we  propose a deep clustering framework, \textbf{R}andom-\textbf{w}alk based \textbf{S}calable \textbf{L}earning (\textbf{RwSL}). To enhance the representative power of input data, we smooth the node attributes with  (a parallelizable implementation of) GPR  \cite{PanLi_2019seed_expansion,cwdlydw2020gbp} serving as a  filter that 
incorporates  the graph topology. We designed a  mechanism to integrate an autoencoder with a Deep Neural Networks (DNN) based co-train module, inspired by 
the recent self-supervised learning paradigm~\cite{self_supervised_CL_Jaiswal_2021,velickovic_2019_DGI,Bo_2020_structural_deep}. 

% Our overall approach involves:  1) learning 
% a powerful embedding from the  encoder and maximizing the information preservation by reducing reconstruction loss from  the decoder; 2) obtaining  clustering assignments distributions from the encoder embedding and the co-train module separately,  and then improving  cluster cohesion by minimizing KL divergence.

To summarize,  we make the following contributions:   
1) We introduce the first deep clustering approach for attributed graphs that  can be scaled to an arbitrary sized  graph, 
% which in turn 
% builds on the derivation of  the relationship between Laplacian smoothing and Generalized PageRank (GPR);  
2) We analyze  spectral properties of a GPR-based graph filter and their influence on clustering performance, 
3) We compare  our new algorithm with a number 
of state-of-the-art graph clustering algorithms, and show how our method achieves 
scalability, while being  competitive (or even outperforming the best known methods) with 
respect to  six clustering metrics.

%% file: text/related_work.tex
% \vspace*{-0.1in}
\vspace*{-2ex} 
\section{Related Work}
\label{sec:related_work}

GAE based methods \cite{kipf2016variational_VGAE,pan_2018_ARGA,wang_mgae_2017} apply  training objectives on reconstructing either adjacency matrix, node attributes, 
or data representation learned by the DNN. Structural Deep Clustering Network (SDCN)~\cite{Bo_2020_structural_deep} combines an autoencoder module for data representation learning
with a GCN module and  trains the two deep neural architectures end-to-end for clustering through a self-supervised mechanism. 
Adaptive Graph Convolution (AGC) \cite{ijcai2019_AGC} is an adaptive graph convolution method for attributed graph clustering that  uses high-order graph convolution to capture global cluster structure. 
Simple Spectral Graph Convolution (SSGC)  \cite{ssgc_2021}  is a variant of GCN that  exploits a modified Markov Diffusion Kernel. Deep Modularity Networks (DMoN) \cite{tsitsulin2020graph_dmon} is based on GCN and is an unsupervised pooling method inspired by the modularity measure of clustering quality. The contrastive self-supervised learning based Deep graph Infomax (DGI)~\cite{velickovic_2019_DGI} incorporates GCN modules and achieves advanced embedding clustering.

%% file: text/methods.tex
\section{Methodology} 
\label{sec:method}

%Our framework is composed of two modules: 1) an adaptive auto-encoder; 2) a co-train self-supervised DNN, i.e. a multi-layer perceptron that  integrates the latent representation from the auto-encoder during the training. 

% \vspace*{-2ex} 
\subsection{Overview}

Consider an  attributed graph  $\mathcal{G} = (\mathcal{V}, \mathcal{E}, X)$, where 
$\mathcal{V}={v_{1}, v_{2}, \cdots, v_{n}}$ denotes a set of $n$ vertices and $d_{j}$ denotes the degree of node $j$.  $\mathcal{E}$ is the edge set and $X \in \mathbb{R}^{n \times d}$ is the feature matrix, where each node $v$ is associated with a $d$-dimensional feature vector $X_{v}$.  The node clustering task aims to partition the  nodes into multiple disjoint groups of 
similar nodes.  

\begin{figure}
% \begin{center} 
  \includegraphics[width=\columnwidth]{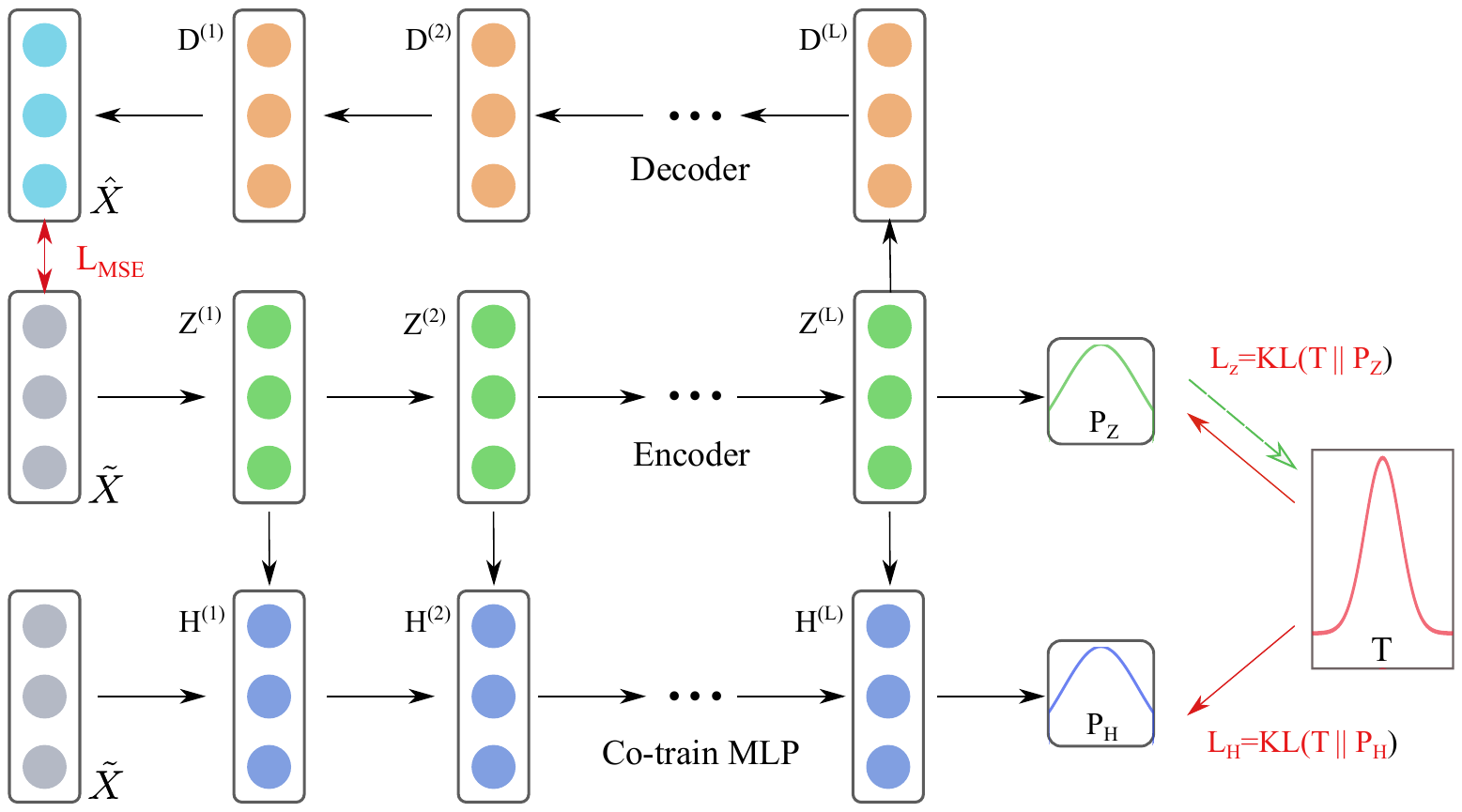}
  \caption{RwSL framework architecture. $\Tilde{X}$ is the input filtered data; $\hat{X}$ is the reconstructed data from the decoder. $Z^{l}$ and $D^{l}$ are the latent representation learned from the encoder and decoder. $H^{l}$ is the latent representation from co-train self-supervised DNN module. $P_{Z}$ and $P_{H}$ are respectively the probability distribution of sample assignment to clusters from the last layer of encoder and co-train MLP. The target distribution $T$,  initially calculated from $P_{Z}$, will guide the optimization of both auto-encoder and co-train DNN modules.}
  \label{fig:arch} 
%   \end{center} 
\end{figure}

As stated previously, Laplacian smoothing filtering has been used in recent 
node clustering efforts~\cite{SGC_2019,ijcai2019_AGC,ssgc_2021}.
The key foundation of our work is the derivation 
of the relationship between Laplacian smoothing and Generalized PageRank (GPR).  
This allows us to perform a novel application of the  graph filtering  previously 
proposed (in a different context)  by \cite{cwdlydw2020gbp}, 
resulting in a  filtered attribute matrix $\Tilde{X}$ that combines 
information from both node features and graph topology. If computation of $\Tilde{X}$
can be performed on large graphs, the computed values can be  fed to DNN based autoencoder. 
In addition, inspired by the self-supervised  learning paradigm, a co-training DNN component is included  to further improve the clustering accuracy and performance.
Resulting overall method  will be presented in 
Section~\ref{subsec:overall} but is summarized in Figure \ref{fig:arch}.

% \vspace*{-2ex} 
\subsection{Laplacian Smoothing Filters and Generalized PageRank} 
\label{subsec:gpr}

% \begin{equation}\label{eq:Laplacian smoothing matrix}
%     \begin{split}
%         \hat{Y} = (I-\gamma\Tilde{D}^{-1}\Tilde{L})X \\
%     \end{split}
% \end{equation}
Let $A$ be the adjacency matrix  
capturing the edges in the graph and $D$ be the diagonal degree matrix. The graph Laplacian matrix is defined as $L=D-A$. Let $\Tilde{A} = A + I_{n}$ denote the augmented adjacency matrix with self-loops added. 
%and the resulting diagonal degree matrix be denoted as 
%$\Tilde{D}$. {\bf Xiang: please check if the symbols are consistent above and below}

The Laplacian smoothing \cite{Taubin_Laplacian_smoothing_1995} can be written in a matrix form as \cite{Li_Deeper_into_GCN_2018}: $\hat{Y} = (I-\gamma\Tilde{D}^{-1}\Tilde{L})X$, where $\Tilde{D}$ and $\Tilde{L}$ ($\Tilde{L}=\Tilde{D}-\Tilde{A}$) are the degree matrix and Laplacian matrix, 
respectively, corresponding to $\Tilde{A}$ defined above. 
If we replace the normalized Laplacian $\Tilde{D}^{-1}\Tilde{L}$ with the symmetric normalized Laplacian $\Tilde{L}_{sym}=\Tilde{D}^{-\frac{1}{2}}\Tilde{L}\Tilde{D}^{-\frac{1}{2}}$, we will obtain the generalized Laplacian smoothing filter $\bf{H}$ \cite{Taubin_Laplacian_smoothing_1995}:

\vspace*{-2ex}  
\begin{equation}
\label{eq:Symmetric Laplacian smoothing matrix}
    \begin{split}
        \hat{Y} = HX = (I-\gamma\Tilde{L}_{sym})X \\
    \end{split}
\end{equation}

GCN is a special form of Laplacian smoothing by setting $\gamma=1$. Previous work~\cite{cwdlydw2020gbp} proposed a scalable approach, GnnBP (Graph neural network via Bidirectional Propagation), to precompute
 the feature propagation with the Generalized PageRank matrix ($P$) \cite{PanLi_2019seed_expansion}. Specifically, in their method:

\vspace*{-2ex}  
\begin{equation}
    \label{eq:pagerank}
    \begin{split}
       P =S \cdot X= \sum\limits_{l=0}^{L}w_{l}T^{l}\cdot X = \sum\limits_{l=0}^{L}w_{l}(\Tilde{D}^{r-1}\Tilde{A}\Tilde{D}^{r})^{l}\cdot X   %\triangleq C^T \cdot X
    \end{split}
\end{equation}

where  $r \in [0, 1]$ is  the convolution coefficient, $w_{l}$'s are  the weights of different order convolution matrices satisfying $\sum^{\infty}_{l=0}w_{l} \leq 1$ and $T^{l}\cdot X=(\Tilde{D}^{r-1}\Tilde{A}\Tilde{D}^{r})^l \cdot X$ is  the $l^{th}$ step propagation matrix (recall that $X$ represents the node attributes). GnnBP \cite{cwdlydw2020gbp} was introduced as  a localized bidirectional propagation algorithm  
for an unbiased estimate of the Generalized PageRank matrix ($P$).  

Now, if we stack  $L$ Laplacian smoothing filters from Eq.\ref{eq:Symmetric Laplacian smoothing matrix}, we obtain: 

\vspace*{-2ex}  
\begin{equation}\label{eq:Laplacian smoothing}
    \begin{split}
      \hat{Y}=H^{L}X
    =(I_{n}-\gamma(I_{n}-\Tilde{D}^{-\frac{1}{2}}\Tilde{A}\Tilde{D}^{-\frac{1}{2}}))^{L}\cdot X \\
       =(\gamma\Tilde{D}^{-\frac{1}{2}}\Tilde{A}\Tilde{D}^{-\frac{1}{2}}+(1-\gamma)I_{n})^{L}\cdot X
    \end{split}
\end{equation}

The expression obtained above  is a special case of the GPR in  Eq.\ref{eq:pagerank}, as can be seen by 
setting $r=0.5$ and manipulating the weights $w_{l}$ to simulate various diffusion processes. %(as also discussed in \cite{klicpera_diffusion_2019}). 
Therefore, the Laplacian smoothing filters as used by prior works \cite{SGC_2019,ijcai2019_AGC,ssgc_2021} can be generalized into GPR -- the key foundation 
of our method and the basis for creating a scalable version. 
% that's correct, SGC and SSGC are filter based methods. Li_2018 is a theory paper, no filters are proposed.

\noindent{\bf Computation:} Considering both clustering performance and scability, we applied bidirection propagation algorithm~\cite{cwdlydw2020gbp} as our precomputation procedure. Specifically, 
we  focused on one setup: $w_l = \alpha(1-\alpha)^{l}$, 
where $\bf{P}$ becomes the PPR (Personalized PageRank) as used in APPNP \cite{Klicpera2019_APPNP} and $S=\sum\limits_{l=0}^{L}w_{l}T^{l}$ corresponds to the PPR diffusion matrix in  a previous 
method~\cite{klicpera_diffusion_2019} ($\alpha$ is the {\em teleport probability}). The size of the neighborhood from which each node will aggregate information can be controlled by $\alpha$, i,e.,  we can balance the needs of preserving locality and leveraging information from a large neighborhood by adjusting $\alpha$.  PPR can even use infinitely many neighborhood aggregation layers without leading to over-smoothing.

% \vspace*{-2ex} 
\subsection{Other Details and Overall Algorithm } 
\label{subsec:overall}

%As data patterns provide the only clues for an unsupervised learning scenario like clustering, therefore how to extract effective information from the attributed graphs will heavily influence the clustering performance.

% With powerful capability of capturing inherent data patterns and learning effective representations, deep neural networks have been successfully   used as the basis for various auto-encoders in recent 
% years\cite{kipf2016variational_VGAE,pan_2018_ARGA,Bo_2020_structural_deep}. 
% These efforts use  reconstruction loss as the main training objective and also 
% use GCNs to  encode the graph structure information. 
% %and are hence suffering from over-smoothing issues and resulting performance degradation.
% In this work, we use GPR \cite{PanLi_2019seed_expansion,cwdlydw2020gbp} based graph filters to absorb the topological information into the filtered attributes as the 
% input to our co-train module  -- thus  avoiding over-smoothing issues and 
% easing  the burden of encoding structural information for 
% auto-encoders.  
\noindent 
{\bf Auto-encoder Module:}
With the smoothed attributes $\Tilde{X}$ from GPR graph filters as input, we employ a multi-layer perceptron in the auto-encoders network, in which the encoder and the decoder are symmetric. The encoder aims to learn a robust embedding $Z^{L}$ and the initial clustering  
is performed using the common K-means algorithm \cite{Kmeans_1979}. Following \cite{AGC_chun_wang_2019,Bo_2020_structural_deep}, we use student's t-distribution  \cite{vanDerMaaten2008} as a kernel to measure the similarity
between  the representation vector $z_i$ (i-th row of $Z^{(L)}$) and the cluster center vector
$\mu_{j}$ for $i^{th}$ sample and $j^{th}$ cluster -- more specifically, 

\vspace*{-2ex} 
\begin{equation}
    \label{eq:cluster freq}
    \begin{split}
    p_{ij} = \frac{(1+\|z_i - \mu_{j} \|^2/v)^{-\frac{v+1}{2}}}{\sum_{j^{'}}(1+\|z_i - \mu_{j^{'}} \|^2/v)^{-\frac{v+1}{2}}} 
    \end{split}
\end{equation}

 where $v$ is the degree of freedom of the Student t-distribution. We obtain  distribution of sample clustering $P_{Z}=[p_{ij}]$ by considering $p_{ij}$ as the probability of assigning the 
 sample $i$ to the  cluster $j$. 
  Based on the intuitive motivation to improve the cluster cohesion, a target distribution $T = [t_{ij}]$ can be calculated upon $P_{Z}$ \cite{Bo_2020_structural_deep}): $t_{ij} = \frac{p_{ij}^{2}/f_{i}}{\sum_{j^{'}}p_{ij^{'}}^2/f_{j^{'}}}$, where, $ f_{j} = \sum_{i}p_{ij}$ are soft cluster frequencies. Therefore, $T$ will guide the entire 
model to improve on clustering performance. The decoder will try to reconstruct the input attributes, yielding $\hat{X}$. The preservation of the input information will be tested by checking the reconstruction loss $L_{MSE} = \frac{1}{2N}{||\Tilde{X}-\hat{X}||^{2}_{F}}$.

% \begin{equation}
%     \label{eq:cluster freq}
%     \begin{split}
%  t_{ij} = \frac{p_{ij}^{2}/f_{i}}{\sum_{j^{'}}p_{ij^{'}}^2/f_{j^{'}}}
%      \end{split}
% \end{equation}

% \begin{equation}\label{eq:MSE loss}
%     \begin{split}
%     L_{MSE} = \frac{1}{2N}{||\Tilde{X}-\hat{X}||^{2}_{F}}
%     \end{split}
% \end{equation}

\noindent 
{\bf Co-train DNN Module:}
The complete training process is shown in Algorithm \ref{alg:RwSL Train}. The purpose of the co-train DNN module is to learn the soft probability of assigning samples to separate clusters $P_{H}$ as from the last DNN layer. 
% To further strengthen the representation learning capability of DNN, we mainly take two precautions: 1) obtain the enhanced input data $\Tilde{X}$ by applying graph filters; 2) integrate the latent representation from auto-encoder $Z^{l}$ into the co-train DNN $H^{l}$ as on line 11.
Inspired by the self-supervised mechanism\cite{Bo_2020_structural_deep}, we optimize the learning parameters from both DNN and auto-encoder modules simultaneously by minimizing the two KL divergence losses:  $L_{Z} = KL(T\|P_{Z}) $ and $L_{H} = KL(T\|P_{H})$, together with the reconstruction loss $L_{MSE}$ as on line 14. 
In the process, we try to preserve the attributed graph information in the auto-encoder embeddings, 
while continuously enforcing  the clusters to be more cohesive. 
% Moreover, the latent representation $H_{l}$ is expected to be helpful towards  clustering  with the topological information absorbed in $\Tilde{X}$ at the beginning of the training. This is because Laplacian smoothed features can ease the clustering tasks 
% \cite{Li_Deeper_into_GCN_2018}. 
The mini-batch training is applied to ensure  a more robust convergence and high scalability.

\begin{small}
\begin{algorithm}[tbh]
	\caption{RwSL Training Process} 
	\label{alg:RwSL Train}
	\begin{itemize}
    \Statex{\textbf{Input data}: filtered attributes $\Tilde{X}$, graph $G$, number of clusters: $K$, maximum iterations \textit{MaxIter}, update period \textit{UpdateIter}}
    
% 	\Statex{\textbf{Output}: cluster assignments $R$; }
    \end{itemize}
	
	\begin{algorithmic}[1]
	    \State{Initialize \textbf{$\mu$} with K-means on the representations learned from pre-trained auto-encoder optimized by $L_{MSE}$;}
	    \For {$iter \in 0,1,\vdots,MaxIter$}
	        \State{Generate encoder representations $Z^{(1)}, Z^{(2)},\vdots, Z^{(L)}$}
	        \If{$iter$ mod $UpdateIter == 0$}
	            \State {Use $Z^{(L)}$ to compute distribution $P_{Z}$} 
	            \State {Calculate target distribution $T$}
	        \EndIf
	        \For{X\_batch in $\Tilde{X}$}
    	        \State{$H^{(0)}=X\_batch$}
    	        \For {$l \in 1, \vdots, L-1$}        	             \State{$H^{(l)}=\phi(\Tilde{H}^{(l-1)}W_{g}^{(l)})$} \State{$\Tilde{H}^{(l)}=(1-\epsilon)H^{(l)}+\epsilon Z^{(l)}$}
    	        \EndFor
    	        \State{$H^{(L)}=\phi(\Tilde{H}^{(L-1)}W_{g}^{(L)})$;} 
        		\State {Feed $Z^{(L)}$ to the decoder to obtain $\hat{X}$;} 
        	    \State{Calculate loss:
        	    $L_{tot} = L_{MSE}+ \beta L_{H} + \gamma L_{Z}$}
        	    \State{Back propagation and update learnable parameters}
        	\EndFor
	    \EndFor
	   % \State{Calculate the cluster assignments: $R \gets argmax(P_{H})$}
	\end{algorithmic} 
\end{algorithm}
\end{small}

% \vspace*{-2ex} 
\subsection{Complexity Analysis}

\begin{table*}
    \centering
    \caption{Time and Space complexity during Training on the GPU O($\cdot$), with $N$ as the number of nodes, $E$ as the number of edges, $F$ as the feature dimension, $L$ as the number of neural network layers, $B$ as the batch size.}
    \begin{adjustbox}{totalheight=\textheight-2\baselineskip, max width=\textwidth}
    \begin{tabular}{l|r|r|r|r|r|r}
    \hline
    Complexity & DGI~\cite{velickovic_2019_DGI} & SDCN \cite{Bo_2020_structural_deep} & DMoN \cite{tsitsulin2020graph_dmon} & AGC \cite{ijcai2019_AGC} & SSGC \cite{ssgc_2021} & RwSL \\
     \hline
    Computation Cost & $L\cdot E \cdot F + L\cdot N \cdot F^{2}$ & $L\cdot E \cdot F + L\cdot N \cdot F^{2}$ & $L \cdot E \cdot F + L \cdot N \cdot F^{2}$ & $E \cdot F + N \cdot F$ & $E\cdot F+N \cdot F$ & $L\cdot B \cdot F^{2}+N\cdot F$ \\
    Memory Cost & $N\cdot F+ E +L\cdot F^{2}$ & $N\cdot F+ E +L\cdot F^{2}$ & $N \cdot F+E  +L\cdot F^{2}$ & $N\cdot F+ E $ & $N\cdot F+ E $ & $B\cdot F+L\cdot F^{2}$ \\
     \hline
    \end{tabular} 
    \end{adjustbox}
    \label{tab: Complexity}
\end{table*}

Table~\ref{tab: Complexity} compares the training complexity among different frameworks. RwSL computational cost includes $L$ layers of matrix multiplication for co-train DNN module: $O(L\cdot B\cdot F^{2})$ and calculating the target distribution: $O(N\cdot F)$. RwSL memory cost consists of saving a batch of input : $O(B\cdot F)$ and storing learning parameters: $O(L\cdot F^{2})$. In contrast, all the other baseline clustering frameworks include an expensive sparse matrix-matrix multiplication of adjacency matrix and feature matrix of the 
entire graph.

%% file: text/analysis.tex
\section{ Analysis and Observations }
\label{sec:theory} 

\begin{figure*}
     \centering
     \begin{subfigure}[b]{0.32\textwidth}
         \centering
         \includegraphics[width=\textwidth]{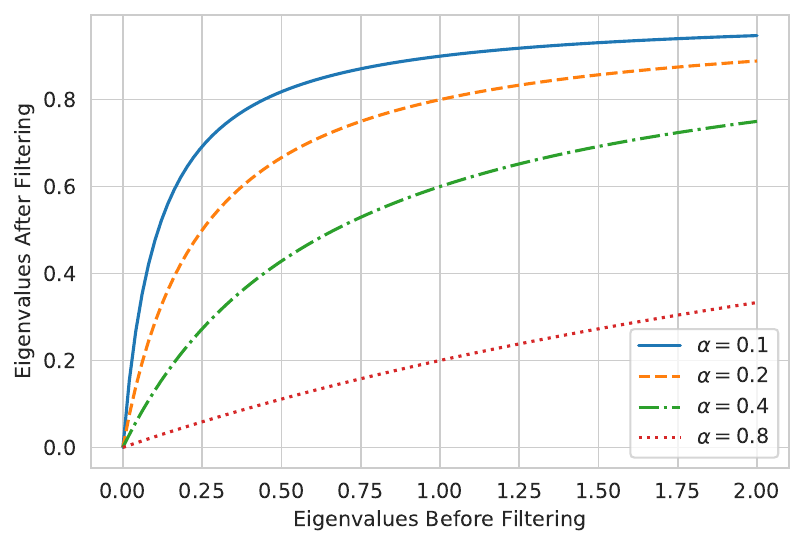}
         \caption{}
         \label{fig:eig filter diff}
     \end{subfigure}
     \hfill
     \begin{subfigure}[b]{0.32\textwidth}
         \centering
         \includegraphics[width=\textwidth]{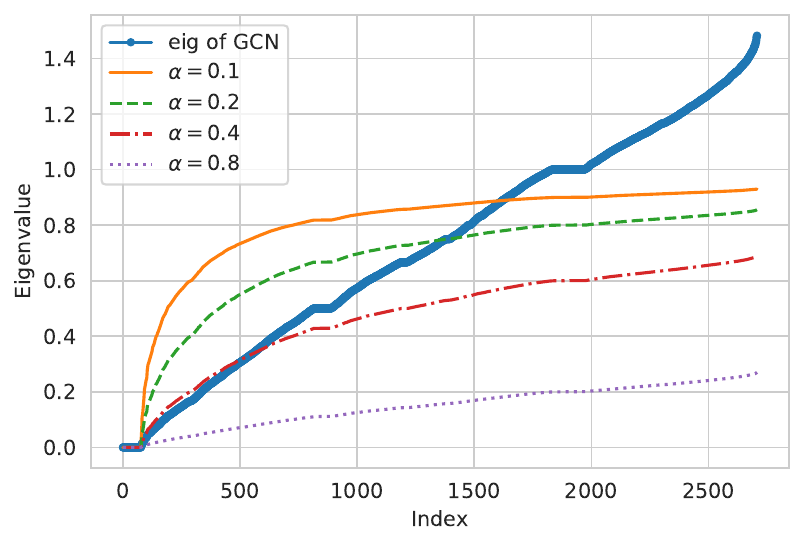}
         \caption{}
         \label{fig:eigenvalues vs index}
     \end{subfigure}
     \hfill
     \begin{subfigure}[b]{0.32\textwidth}
         \centering
         \includegraphics[width=\textwidth]{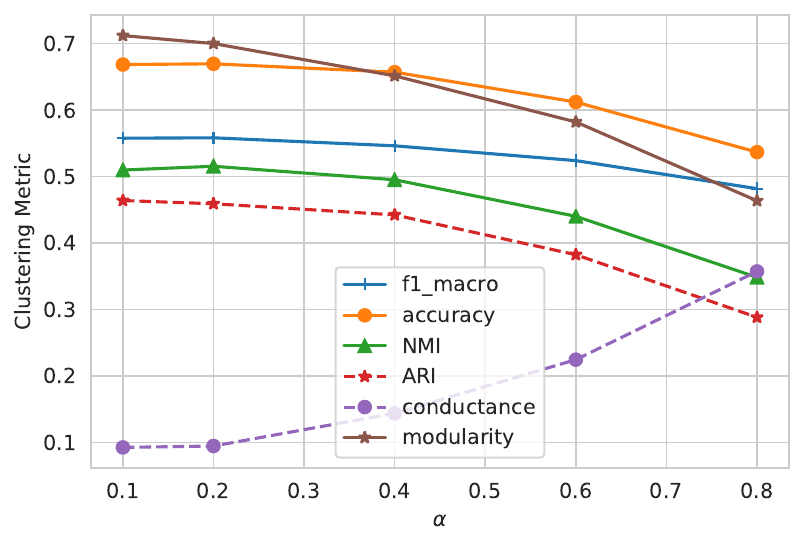}
         \caption{}
         \label{fig:Teleport probability}
     \end{subfigure}
    \caption{(a) Laplacian eigenvalues of PPR filters $\lambda^{\Tilde{L}_{PPR}}_i=1 - \frac{\alpha}{1-(1-\alpha)(1-\lambda_{i})}$ with spectrum $\lambda_{i} \in [0, 2)$; (b) Sorted by index, Laplacian eigenvalues $\lambda^{\Tilde{L}_{sym}}_{i}$ of a GCN filter and Laplacian eigenvalues $\lambda^{\Tilde{L}_{PPR}}_i$ of PPR filters  with multiple $\alpha$ on Cora; (c) Teleport probability $\alpha$ influence on clustering metrics for Cora}
    \label{fig:Cora eigen}
\end{figure*} 

For  a clustering task, it is desirable  that nodes of the same class have similar feature representations after graph filtering, and we analyze the proposed approach from that 
angle. As a background~\cite{ijcai2019_AGC,Spectral_graph_theory_Chung_1997}, each column of a node attribute  can be considered as a {\em graph signal}, which, in turn, 
can be considered {\em smooth}  if nodes in the neighborhood have similar representations. 
Furthermore, smooth graph signal should contain more low-frequency than high-frequency basis signals \cite{ijcai2019_AGC}.

We denote  $\lambda^{\Tilde{L}_{sym}}_{i}$ as the eigenvalue of the Laplacian $\Tilde{L}_{sym} = I-\Tilde{D}^{-\frac{1}{2}}\Tilde{A}\Tilde{D}^{-\frac{1}{2}}$. 
Similarly,  the eigenvalue  of the GCN feature propagation  $T=\Tilde{D}^{-\frac{1}{2}}\Tilde{A}\Tilde{D}^{-\frac{1}{2}}$ is denoted as $\Tilde{\lambda}^{GCN}_{i} = 1-\lambda^{\Tilde{L}_{sym}}_{i}$ based on $\bf{L}=\bf{I}-\bf{T}$. A K-layer GCN  corresponds to filters with the frequency response function  $g(\lambda^{\Tilde{L}_{sym}}_{i})=(1-\lambda^{\Tilde{L}_{sym}}_{i})^{K}$ and acts as a low-pass-type filter by taking $K>1$~\cite{SGC_2019}.  

Before presenting our analysis, we consider the following background. 
It is known that low and high frequencies can capture global and local contexts, 
respectively, of each node\cite{ssgc_2021}. However, repeatedly applying Laplacian smoothing can lead to {\em over-smoothing},  where features of nodes within each connected component of the graph will converge to (almost) identical values \cite{Li_Deeper_into_GCN_2018}.  
Similarly, increasing number of propagation layers can also 
lead to the filter losing local information associated with the nodes~\cite{Jump_Xu_2018,Klicpera2019_APPNP}. 

The PPR based propagation scheme \cite{Klicpera2019_APPNP} ensures the PageRank score encodes the local neighborhood of root nodes \cite{Page1999ThePC} through a controllable 
parameter {\em teleporting probability} ($\alpha$). 
%Since eigenvectors won't be affected by summation and PPR could include infinite neighbor nodes without over-smoothing \cite{Klicpera2019_APPNP}, 
We investigate the eigenvalues ($\Tilde{\lambda}_{i}^{PPR}$) for the PPR based graph filter with infinite propagation layer (denoted as $ \Tilde{\lambda}_{i}^{PPR^{\infty}}$  after 
setting $L \to \infty$ in eq.\ref{eq:pagerank})  which can be transformed to a closed form \cite{klicpera_diffusion_2019}:

\vspace*{-2ex}  
\begin{equation}
    \label{eq:eigenvalue}
    % \begin{split}
 \Tilde{\lambda}_{i}^{PPR^{\infty}} =  \alpha\sum\limits_{l=0}^{\infty}((1-\alpha)\Tilde{\lambda}_{i}^{GCN})^{l}
 = \frac{\alpha}{1-(1-\alpha)\Tilde{\lambda}_i^{GCN}}
    % \end{split}
\end{equation}

Based on Eq.~\ref{eq:eigenvalue}, we develop the following claims:

\begin{claim}
\label{receptive_claim}
In the PPR graph filter with infinite propagation, $ S= \sum\limits_{l=0}^{\infty}w_{l}(\Tilde{D}^{-\frac{1}{2}}\Tilde{A}\Tilde{D}^{-\frac{1}{2}}))^{l} $ where $w_{l}=\alpha(1-\alpha)^{l}$, we have: for $0 < \alpha_{1} < \alpha_{2} < 1$, $\exists  l_{0}$ such that $\forall l > l_{0}$, $w_{l}(\alpha_{1}) > w_{l}(\alpha_{2})$.  
\end{claim}

\begin{proof}

The weight of each l-hop propagation $w(l)=\alpha(1-\alpha)^{l}$ for $\alpha \in (0, 1)$ has the following properties: 

\begin{enumerate} 
\item  $\sum\limits_{l=0}^{\infty}w_{l}=\sum\limits_{l=0}^{\infty}\alpha(1-\alpha)^{l}=1$, 
\item  $w(l)$ monotonically decreases with increasing $l$.  
\end{enumerate} 

\textbf{Step 1}, we first prove the following fact: 
$\exists  l_{0} > 0 $ such that $\alpha_{1}(1-\alpha_{1})^{l_0} \ge \alpha_{2}(1-\alpha_{2})^{l_0} $ for $ 0 < \alpha_{1} < \alpha_{2} < 1 $. 

 This can be proved by contradiction. Assume, $\forall l > 0$, we have $\alpha_{1}(1-\alpha_{1})^{l} < \alpha_{2}(1-\alpha_{2})^{l} $ for $ 0 < \alpha_{1} < \alpha_{2} < 1 $. Then $\sum\limits_{l=0}^{\infty}\alpha_{1}(1-\alpha_{1})^{l} < \sum\limits_{l=0}^{\infty}\alpha_{2}(1-\alpha_{2})^{l} = 1$. This contradicts the fact that $\sum\limits_{l=0}^{\infty}\alpha_{1}(1-\alpha_{1})^{l} = 1$. 
 
\textbf{Step 2}, $\forall l_{1} > l_{0}$, we have
\begin{equation*}\label{eq:weight proof}
    \begin{split}
        \alpha_{1}(1-\alpha_{1})^{l_{1}} 
        = \alpha_{1}(1-\alpha_{1})^{l_{0}}(1-\alpha_{1})^{l_{1} - l_{0}} \\
    \ge \alpha_{2}(1-\alpha_{2})^{l_0}(1-\alpha_{1})^{l_{1} - l_{0}} \\
    > \alpha_{2}(1-\alpha_{2})^{l_0}(1-\alpha_{2})^{l_{1} - l_{0}} = \alpha_{2}(1-\alpha_{2})^{l_{1}} \\
    \end{split}
\end{equation*}

\end{proof}

\begin{figure}
     \centering
     \includegraphics[width=\linewidth]{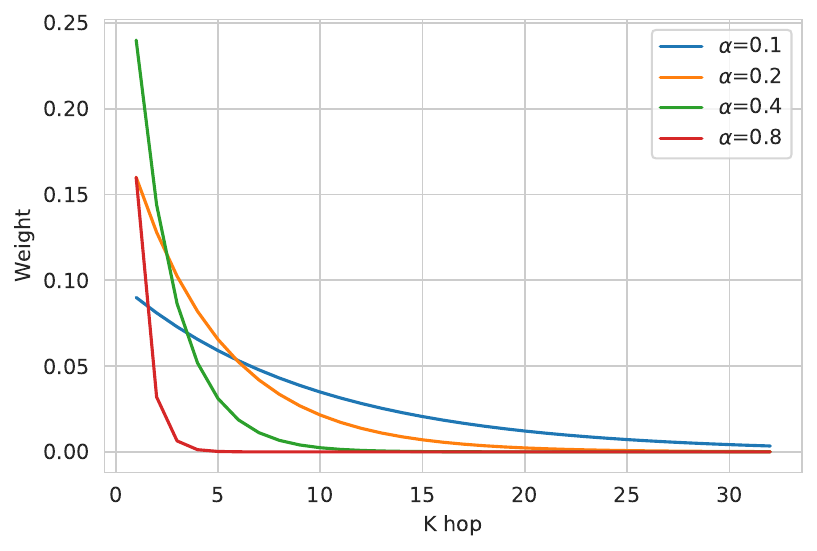}
     \caption{PPR filter weights $w_{l}$ for different hops of propagation}
     \label{fig:weight alpha}
\end{figure}

As indicated by $w(l)=\alpha(1-\alpha)^{l}$ with $\alpha \in (0, 1)$, the weights decrease exponentially with the propagation distance. A smaller value $\alpha$ will push the distribution of $w(l)$ to weigh more on larger $l$ as in Figure \ref{fig:weight alpha}.

Thus, a smaller teleport probability $\alpha \in (0, 1)$ indicates a larger receptive field (largest $L$ where $w_{L} > \epsilon_{0}$)  (for a threshold $\epsilon_{0}>0$), covering L-hop of neighborhood.

\begin{claim}
\label{freq_claim}
If we  denote Laplacian eigenvalues of PPR graph filter with infinite propagation as $ \lambda^{\Tilde{L}_{PPR}}_i(\alpha) =1 - \Tilde{\lambda}_{i}^{PPR^{\infty}}(\alpha) = 1 - \frac{\alpha}{1-(1-\alpha)(1-\lambda)} $, where $\lambda$ is a Laplacian eigenvalue, we have
 
 $\forall \lambda \in (0, 2)$, for $0 < \alpha_{1} < \alpha_{2} < 1, $,  $\lambda^{\Tilde{L}_{PPR}}_i(\alpha_{1}) > \lambda^{\Tilde{L}_{PPR}}_i(\alpha_{2}) $
 
\end{claim}

\begin{proof}
With $\lambda \in (0, 2)$ as a constant value, we take the partial derivative of $\lambda^{\Tilde{L}_{PPR}}_i(\alpha)$ regarding $\alpha \in (0, 1)$

\[ \frac{\partial\lambda^{\Tilde{L}_{PPR}}_i(\alpha)}{\partial \alpha} = -\frac{\lambda}{((1-\alpha)\lambda+\alpha)^{2}} < 0 \]

Therefore, for $0 < \alpha_{1} < \alpha_{2} < 1$, we have $\lambda^{\Tilde{L}_{PPR}}_i(\alpha_{1}) > \lambda^{\Tilde{L}_{PPR}}_i(\alpha_{2}) $

\end{proof}

Overall, the significance here is that a smaller teleport probability can include information of nodes from further distance  (Claim~\ref{receptive_claim}), while maintaining node locality by capturing the  high frequency  signals (Claim~\ref{freq_claim}).
%Previous work~\cite{klicpera_diffusion_2019} demonstrated that $\bf{S}$ is a dense matrix even with summation to finite hops of propagation and used a threshold $\epsilon$ to truncate small values of $\bf{S}$ where typical thresholds have negligible effects on eigenvalues, especially those helpful for graph learning. GnnBP \cite{cwdlydw2020gbp} made a trade-off between efficiency and accuracy by adjusting the push threshold $r_{max}$.

%% Xiang: I am not sure what the last 2-3 sentences are connected to the rest 
%% of this narrative ... do your 

%% Reply: our claims are all talking about the infinite case, since only infinity can yield a closed form, which can be feasibly discussed and used to generate figures as in Figure 2a and Figure 2b. These 2-3 sentences show people our discussions could extend to real implementation when using a threshold. Previous work already shows that finite summation with a truncate can barely affect the eigenvalues. This also gives reviewers a explanation how we simulate/approximate filters in the GnnBP implementation by controlling thresholds.

We empirically evaluate  the significance of our observations.
Figure~\ref{fig:eig filter diff} presents Laplacian eigenvalues of the PPR filter in a typical spectrum range. Figure~\ref{fig:eigenvalues vs index} shows Laplacian eigenvalues of a single GCN layer filter compared with those of PPR filters with multiple $\alpha$ values on the Cora graph. We can observe that the PPR graph filter can preserve the low-frequency of Laplacian and a smaller $\alpha$ can retain more high frequency signals.  
%% add a sentence on how this adds to your claims or the practical implication 
%% of this statement 
%% needs explanation of what this shows, relationship to theory, and practical 
%% significance 
%% Reply: both Figure 2a and 2b show the same thing, i.e. the low-pass properties of such filter. Figure 2a shows a filtering effect for all valid Laplacian eigenvalue range [0, 2), while Figure 2b just show the effects on filtering a practical data to align with clustering performance
Finally, we evaluate the influence of $\alpha$ on the clustering performance for Cora   in Figure~\ref{fig:Teleport probability}. The inclusion of high-frequency by using small $\alpha$ can capture more information about local neighborhood of each node and yields 
better performance with respect to clustering metrics. As  $w(l)=\alpha(1-\alpha)^{l}$ with $\alpha \in (0, 1)$ decreases exponentially with the propagation distance $l$, the PPR filter maintains node locality by weighing more on the nodes in closer neighborhood and thus relieves the over-smoothing issues when including nodes from larger neighborhood.

%% file: text/experiment.tex
\section{Experimental Results}  
\label{sec:expr}  
% \vspace*{-2ex} 

% \begin{wraptable}{r}{3.0in} 
\begin{small} 
\begin{table}[tbh]
    % \centering
    \caption{Datasets Statistics}
    \begin{adjustbox}{max width=\columnwidth,center}
    \begin{tabular}{lrrrr}
    \hline
    dataset  & Nodes & Classes & Features & Edges \\ 
    \hline
    ACM       & $ 3025  $  & $ 3 $  & $ 1870 $   & $13,128$ \\
    DBLP      & $ 4058 $   & $ 4 $  & $ 334 $    & $3528$ \\
    Citeseer  & $ 3327  $  & $ 6 $  & $ 3703 $   & $4732$ \\
    Cora      & $ 2708  $  & $ 7 $  & $ 1433 $   & $5429$ \\
 %   Pubmed    & $ 19717  $  & $ 3 $  & $ 500 $   & $44,338$ \\
 %   Amazon Photo  & $ 7650 $   & $ 8 $ & $ 745 $    &  $71,831$ \\
    Coauthor CS     & $ 18333 $  & $ 15 $ & $ 6805 $  &  $81,894$   \\
    Coauthor PHY     & $ 34493 $  & $ 5 $ & $ 8415 $  &  $247,962$  \\
    Friendster   &  $6.5 \times  10^7$   & $ 7 $ & $ 40 $  &  $1.8 \times 10^9$  \\
    \hline
    \end{tabular}
    \end{adjustbox}
    \label{tab:Datasets info}
\end{table}   
\end{small}  
% \end{wraptable} 

\vspace*{-2ex} 
\subsection{Datasets}
The RwSL is evaluated and compared with existing state-of-the-art on 6/7   real-world datasets  that 
are summarized in Table \ref{tab:Datasets info} (the seventh dataset, Friendster, could not be used on 
other methods because of these methods'  scalability limitations): 
\textbf{ACM} \cite{Bo_2020_structural_deep}, \textbf{DBLP} \cite{Bo_2020_structural_deep},  {\bf  Citeseer}  and {\bf Cora} \cite{Sen_2008_citation_networks},  {\bf Coauthor CS}  and {\bf Coauthor PHY} \cite{shchur2019pitfalls}, \textbf{Friendster} \cite{yang_friendster_2012}.

\vspace*{-2ex} 
\subsection{Experimental Settings} 

\noindent 
{\bf Baselines:}
We compared the proposed RwSL model with several other (graph) clustering frameworks.  \textbf{KMeans}~\cite{Kmeans_1979}, a classical clustering method that uses 
node attributes only;  
\textbf{DeepWalk} \cite{Perozzi_deepwalk_2014}, an approach for learning latent representations of vertices in a graph using topological information only. Next, we 
used five most advanced deep clustering frameworks as introduced in Section \ref{sec:related_work}: \textbf{DGI}~\cite{velickovic_2019_DGI}, \textbf{AGC}~\cite{ijcai2019_AGC}, \textbf{SDCN }~\cite{Bo_2020_structural_deep}, \textbf{DMoN} \cite{tsitsulin2020graph_dmon} \textbf{SSGC }~\cite{ssgc_2021}.
   
%  In justifying our choice of baselines, we observe that 
%   attributed graph clustering methods,  which utilize both node features and graph structure,  achieve a significant clustering improvement over other approaches that 
%   only exploit one of them. Earlier attributed graph clustering frameworks such as \textbf{GAE} and 
%   \textbf{VGAE}\cite{kipf2016variational_VGAE}, \textbf{ARGE}  and 
%   \textbf{ARVGE}\cite{pan_2018_ARGA} were outperformed by either or both of \textbf{AGC} and \textbf{SSGC}. Deep Embedded Clustering (\textbf{DEC}) \cite{Xie_DEC_2016}, Improved Deep Embedded Clustering (\textbf{IDEC})\cite{ijcai_IDEC_2017}, \textbf{MGAE}\cite{wang_mgae_2017} and Deep Attentional Embedded Graph Clustering (\textbf{DAEGC})\cite{wang_DAEGC_2019} were outperformed by \textbf{SDCN} as baselines, and 
%   finally, \textbf{DMoN} was shown to outperform \textbf{MinCut} \cite{bianchi_MinCut_2020}.  

\noindent 
\textbf{Metrics:} We employ six popular metrics: accuracy, Normalized Mutual Information (NMI), Average Rand Index (ARI), and  macro F1-score are four metrics for ground-truth label analysis, 
whereas modularity~\cite{newman_2006_modularity} and conductance~\cite{yang_friendster_2012} are graph-level metrics. All metrics except conductance will indicate a better clustering with a larger value. 

% Besides accuracy, NMI and macro F1-score, which were generally used by all previous 
% studies, we included modularity and conductance as used in the  DMoN \cite{tsitsulin2020graph_dmon} and ARI as used in the SDCN \cite{Bo_2020_structural_deep}. 

% \noindent 
% \textbf{Parameter Settings:} Our RwSL framework is implemented in PyTorch 1.7 on CUDA 10.1, whereas   the graph filtering procedure is in C++. Our experiments are performed on nodes with a dual Intel Xeon 6148s @2.4GHz CPU and dual NVIDIA Volta V100 w/ 16GB memory GPU and 384 GB DDR4 memory. 
% Graph filtering is executed on CPU while both autoencoder pretraining and DNN co-train process are performed on a single GPU. We used dropout to make our model more robust and applied a AdamW optimization method with a decoupled weight decay regularization technique \cite{loshchilov2019decoupled_weight_decay}. 

\begin{table*}
    \centering
    \textbf{\caption{Clustering performance on six datasets (mean$\pm$std). RwSL or RwSL-R results  highlighted in bold if they have the 
    top 2 clustering performance.   The asterisk indicates a convergence issue.\label{tab:Results-of-clustering-metrics}}}
    \centering
    \begin{adjustbox}{totalheight=\textheight-2\baselineskip, max width=\textwidth}
    \begin{tabular}{lrrrrrr} 
    
    \hline
    Dataset & Accuracy $\uparrow$ & NMI $\uparrow$ & ARI $\uparrow$ & macro\_F1 $\uparrow$ & Modularity $\uparrow$ & Conductance $\downarrow$ \\
    \hline
    \textbf{ACM} \\
    KMeans & $ 66.62 \pm 0.55 $  & $ 32.41 \pm 0.34 $  & $ 30.22 \pm 0.41 $ & $ 66.83 \pm 0.57 $  & $31.20 \pm 0.50 $ & $ 30.96 \pm 0.23 $  \\
    % filter-Kmeans & $ 65.52 \pm 0.22 $  & $ 32.28 \pm 0.31 $  & $ 26.70 \pm 0.31 $ & $ 66.55 \pm 0.22 $  & $42.15 \pm 0.11 $ & $ 17.33 \pm 0.25 $  \\
    DeepWalk & $ 50.59 \pm 4.27 $  & $ 16.12 \pm 4.96 $  & $ 18.56 \pm 5.80 $ & $ 46.56 \pm 4.43 $  & $ 38.57 \pm 9.51 $ & $ 1.79 \pm 0.59 $  \\
    SDCN  & $ 89.63 \pm 0.31 $  & $ 66.74 \pm 0.75 $  & $ 72.00 \pm 0.75 $ & $ 89.60 \pm 0.32 $  & $60.86 \pm 0.16 $ & $ 3.07 \pm 0.17 $  \\
    DMoN   & $ 88.51 \pm 0.72 $  & $ 63.59 \pm 1.49 $  & $ 69.04 \pm 1.70 $ & $ 88.57 \pm 0.72 $  & $ 61.57 \pm 0.17 $  & $ 3.16 \pm 0.09 $  \\
    AGC   & $ 78.21 \pm 0.00 $  & $ 46.31 \pm 0.01 $  & $ 48.02 \pm 0.00 $ & $ 78.26 \pm 0.00 $  & $ 59.44 \pm 0.02 $  & $ 2.51 \pm 0.01 $  \\
    SSGC   & $ 84.43 \pm 0.29 $  & $ 56.15 \pm 0.51 $  & $ 60.17 \pm 0.60 $ & $ 84.44 \pm 0.29 $  & $ 60.19 \pm 0.05 $  & $ 2.54 \pm 0.11 $  \\
    DGI   & $ 90.17 \pm 0.28 $  & $ 67.84 \pm 0.72 $  & $ 73.28 \pm 0.66 $ & $ 90.12 \pm 0.27 $  & $ 59.79 \pm 0.19 $  & $ 3.87 \pm 0.14 $  \\
    RwSL   & $\textbf{ 90.68}  \pm 0.06 $  & $ \textbf{69.08} \pm 0.08 $  & $ {\bf 74.54}  \pm 0.12 $ & $ {\bf 90.65}  \pm 0.06 $  & $ {\bf 61.31}  \pm 0.03 $  & $ 2.99 \pm 0.07 $  \\
    RwSL-R & $ \textbf{90.86} \pm 0.25 $  & $ \textbf{68.79}  \pm 0.43 $  & $ \textbf{74.85} \pm 0.59 $ & $ \textbf{90.84} \pm 0.27 $  & $61.29 \pm 0.04 $  & $ 3.32 \pm 0.21 $  \\
    \hline
    \textbf{Cora} \\
    KMeans & $ 35.37 \pm 3.72 $  & $ 16.64 \pm 4.21 $  & $ 9.31 \pm 2.14 $ & $ 31.49 \pm 4.58 $  & $ 20.77 \pm 3.37 $ & $ 59.77 \pm 5.31 $  \\
    % filter-Kmeans & $ 59.78 \pm 2.71 $  & $ 48.76 \pm 1.60 $  & $ 36.85 \pm 2.28 $ & $ 50.79 \pm 2.05 $  & $64.90 \pm 1.67 $ & $ 12.21 \pm 2.97 $  \\
    DeepWalk & $ 63.87 \pm 2.14 $  & $ 44.11 \pm 1.33 $  & $ 39.64 \pm 1.68 $ & $ 57.98 \pm 2.43 $  & $ 72.98 \pm 0.79 $ & $ 7.88 \pm 0.35 $  \\
    SDCN  & $ 64.27 \pm 4.87 $ & $ 47.39 \pm 3.49 $  & $ 39.72 \pm 5.53 $ & $ 57.88 \pm 6.99 $  & $62.59 \pm 5.18 $  & $ 18.32 \pm 2.26$  \\
    DMoN & $ 58.77 \pm 6.45 $ & $ 46.10 \pm 4.12 $  & $ 37.42 \pm 5.27 $ & $ 46.19 \pm 7.47 $  & $ 68.70 \pm 1.94 $  & $ 8.28 \pm 0.74 $  \\
    AGC   & $ 65.23 \pm 0.93 $  & $ 50.05 \pm 0.49 $  & $ 40.23 \pm 0.95 $ & $ 58.93 \pm 1.68 $  & $ 69.98 \pm 0.46 $  & $ 11.08 \pm 1.61 $  \\
    SSGC & $ 68.50 \pm 1.98 $ & $ 52.80 \pm 1.03 $  & $ 45.70 \pm 1.28 $ & $ 64.38  \pm 2.71 $  & $ 73.71 \pm 0.45 $  & $ 9.41 \pm 0.55 $  \\
    DGI   & $ 68.47 \pm 1.43 $  & $ 52.60 \pm 0.88 $  & $ 45.63 \pm 1.44 $ & $ 65.79 \pm 1.53 $  & $ 69.86 \pm 0.29 $  & $ 13.64 \pm 0.69 $  \\
    RwSL & $ 66.49 \pm 0.99 $ & $ 50.90 \pm 0.51 $  & $ 45.64 \pm 1.91 $ & $ 55.57 \pm 0.65 $  & 
     $ 71.09 \pm 0.14 $ &  $ \textbf{9.17} \pm 0.81 $  \\
     RwSL-R & $ \textbf{70.74} \pm 0.60 $ & ${\bf  52.69}  \pm 0.37 $  & $ \textbf{47.43} \pm 1.06 $ & $ {\bf 62.87}  \pm 0.46 $  & 
     $ {\bf 72.99}  \pm 0.35 $ &  $ 9.33 \pm 0.34 $  \\
    \hline
    \textbf{Citeseer} \\
    KMeans & $ 46.70 \pm 4.33 $  & $ 18.42 \pm 3.26 $  & $ 18.42 \pm 3.26 $ & $ 44.47 \pm 4.44 $  & $43.57 \pm 2.67 $ & $ 37.21 \pm 2.19 $  \\
    % filter-Kmeans & $ 65.51 \pm 6.26 $  & $ 41.97 \pm 3.33 $  & $ 41.60 \pm 5.01 $ & $ 57.13 \pm 6.33 $  & $73.91 \pm 2.35 $ & $ 2.90 \pm 0.62 $  \\
    DeepWalk & $ 43.56 \pm 1.03 $  & $ 16.02 \pm 0.56 $  & $ 16.37 \pm 0.66 $ & $ 40.37 \pm 0.97 $  & $ 76.44 \pm 0.20 $ & $ 2.98 \pm 0.12 $  \\
    SDCN  & $ 63.42 \pm 3.31 $ & $ 37.28 \pm 2.19 $ & $ 37.40 \pm 2.79 $  & $ 56.16 \pm 4.53 $  & $70.83 \pm 2.77$  & $7.98 \pm 1.99 $  \\
    DMoN & $ 47.60 \pm 3.88 $ & $ 24.86 \pm 3.00 $ & $ 22.06 \pm 3.12 $   & $ 45.52 \pm 3.78 $  & $ 77.06 \pm 0.44 $  & $ 5.28 \pm 0.66 $  \\
    AGC   & $ 67.18 \pm 0.52 $  & $ 41.37 \pm 0.70 $  & $ 42.10 \pm 0.87 $ & $ 62.68 \pm 0.48 $  & $ 77.57 \pm 0.21 $  & $ 1.72 \pm 0.04 $  \\
    SSGC & $ 67.86 \pm 0.26 $ & $ 41.86 \pm 0.22 $ & $ 42.95 \pm 0.30 $   & $ 63.61 \pm 0.23 $  & $ 78.03 \pm 0.12 $  & $ 1.75 \pm 0.03 $  \\
    DGI   & $ 68.68 \pm 0.76 $  & $ 43.22 \pm 0.91 $  & $ 44.53 \pm 1.02 $ & $ 64.41 \pm 0.70 $  & $ 72.42 \pm 0.38 $  & $ 7.19 \pm 0.55 $  \\
    RwSL & $ \textbf{70.22} \pm 0.07 $ & $ {\bf 44.30} \pm 0.15 $ & $ \textbf{46.07} \pm 0.16 $   & $ \textbf{66.10} \pm 0.09 $  & $ \textbf{78.53} \pm 0.04 $ & $ 1.89 \pm 0.03 $  \\
    RwSL-R & $ {\bf  69.37}  \pm 0.18 $ & $ \textbf{44.42} \pm 0.20 $ & $ {\bf 45.87}  \pm 0.23 $   & $ {\bf 65.13}  \pm 0.18 $  & $ {\bf 78.28}  \pm 0.07 $ & $ 1.99 \pm 0.02 $  \\
    \hline
    \textbf{DBLP} \\
    KMeans & $ 38.65 \pm 0.58 $  & $ 11.56 \pm 0.53 $  & $ 6.95 \pm 0.39 $ & $ 31.81 \pm 0.53 $  & $33.83 \pm 0.47 $ & $ 36.20 \pm 0.51 $  \\
    % filter-Kmeans & $ 50.23 \pm 0.30 $  & $ 24.00 \pm 0.14 $  & $ 16.13 \pm 0.18 $ & $ 50.80 \pm 0.33 $  & $61.44 \pm 0.04 $ & $ 9.33 \pm 0.08 $  \\
    DeepWalk & $ 38.99 \pm 0.02 $  & $ 5.91 \pm 0.02 $  & $ 5.83 \pm 0.02 $ & $ 36.87 \pm 0.02 $  & $ 64.05 \pm 0.03 $ & $ 4.03 \pm 0.02 $  \\
    SDCN  & $ 69.08 \pm 1.95 $ & $ 34.64 \pm 1.94 $  & $ 36.31 \pm 2.86 $ & $ 67.81 \pm 3.46 $  & $63.38 \pm 1.87$  & $7.56 \pm 0.54$  \\
    DMoN & $ 62.11 \pm 6.53 $ & $ 32.54 \pm 5.05 $  & $ 33.37 \pm 6.43 $ & $ 61.10 \pm 6.82 $  & $ 65.63 \pm 0.66 $  & $ 6.47 \pm 0.54 $  \\
    AGC   & $ 69.06 \pm 0.06 $  & $ 37.00 \pm 0.07 $  & $ 33.69 \pm 0.13 $ & $ 68.59 \pm 0.05 $  & $ 68.77 \pm 0.01 $  & $ 5.29 \pm 0.01 $  \\
    SSGC & $ 68.66 \pm 1.95 $ & $ 33.89 \pm 2.08 $  & $ 37.30 \pm 3.13 $ & $ 65.91 \pm 2.19 $  & $ 62.02 \pm 1.64 $  & $ 3.24 \pm 0.52 $  \\
    DGI   & $ 59.72 \pm 4.68 $  & $ 26.90 \pm 4.43 $  & $ 25.12 \pm 4.76 $ & $ 59.31 \pm 4.69 $  & $ 50.16 \pm 3.77 $  & $ 13.84 \pm 1.12 $  \\
    RwSL & $ 68.25 \pm 0.49 $ & $ 34.39 \pm 0.44 $  & $ 34.51 \pm 0.76 $ & $ 68.15 \pm 0.45 $  & 
     $ {\bf 68.36}  \pm 0.28 $ &  $ {4.13}  \pm 0.19 $  \\
    RwSL-R & $ \textbf{70.84} \pm 0.97 $ & $ \textbf{38.47} \pm 0.95 $  & $ \textbf{39.63} \pm 1.51 $ & $ \textbf{70.75} \pm 0.88 $  & 
     $ 68.34 \pm 0.24 $ &  $ 4.95 \pm 0.07 $  \\
     \hline
     \textbf{Coauthor CS} \\
    KMeans & $ 27.96 \pm 1.09 $  & $ 15.42 \pm 2.25 $  & $ 1.02 \pm 0.74 $ & $ 11.68 \pm 1.56 $  & $ 9.61 \pm 1.88 $ & $ 37.12 \pm 4.10 $  \\
    % filter-Kmeans & $ 49.90 \pm 2.01 $  & $ 57.23 \pm 0.46 $  & $ 32.68 \pm 1.76 $ & $ 36.60 \pm 1.97 $  & $47.82 \pm 0.43 $ & $ 42.73 \pm 0.35 $  \\
    DeepWalk & $ 67.10 \pm 2.98 $  & $ 66.67 \pm 0.86 $  & $ 53.66 \pm 2.91 $ & $ 63.36 \pm 2.84 $  & $ 72.88 \pm 0.41 $ & $ 17.09 \pm 0.66 $  \\
    SDCN  & $ 56.86 \pm 3.40 $ & $ 54.79 \pm 2.44 $  & $ 40.41 \pm 4.52 $ & $ 29.36 \pm 3.22 $  & $53.05 \pm 2.02$  & *$23.09 \pm 1.89 $  \\  
    DMoN & $ 64.95 \pm 3.18 $ & $ 70.06 \pm 1.18 $  & $ 60.97 \pm 3.23 $ & $ 45.12 \pm 2.95 $  & $ 70.79 \pm 0.51 $  & $ 15.66 \pm 0.71 $  \\
    AGC   & $ 62.24 \pm 1.81 $  & $ 65.22 \pm 0.44 $  & $ 46.96 \pm 3.54 $ & $ 51.42 \pm 1.27 $  & $ 69.58 \pm 0.14 $  & $ 19.80 \pm 0.24 $  \\
    SSGC & $ 66.19 \pm 1.19 $ & $ 70.06 \pm 0.67 $  & $ 58.50 \pm 0.17 $ & $ 60.17 \pm 1.94 $  & $ 71.82 \pm 0.14 $  & $ 19.76 \pm 0.22 $  \\
    RwSL & $ \textbf{71.87} \pm 3.16 $ & $ 68.44 \pm 2.31 $  & $ \textbf{64.73} \pm 6.26 $ & $ 49.19 \pm 3.03 $  & 
     $ 66.62 \pm 1.74 $ &  $ 20.88 \pm 2.05 $  \\
    RwSL-R & $ \textbf{68.05} \pm 1.70 $ & $ \textbf{73.77} \pm 0.55 $  & $ {\bf 62.15}  \pm 0.97 $ & $ 57.68 \pm 2.01 $  & 
     $ 71.02 \pm 0.47 $ &  $ 20.26 \pm 0.55 $  \\
    \hline
     \textbf{Coauthor PHY} \\
    KMeans & $ 56.19 \pm 0.75 $  & $ 11.72 \pm 1.92 $  & $ 8.25 \pm 1.26 $ & $ 24.74 \pm 2.11 $  & $ 5.74 \pm 0.83 $ & $ 10.56 \pm 1.47 $  \\
    % filter-Kmeans & $ 61.72 \pm 0.05 $  & $ 56.49 \pm 0.02 $  & $ 40.75 \pm 0.02 $ & $ 65.55 \pm 0.03 $  & $47.49 \pm 0.00 $ & $ 21.47 \pm 0.03 $  \\
    DeepWalk & $ 87.97 \pm 0.01 $  & $ 69.13 \pm 0.02 $  & $ 79.15 \pm 0.03 $ & $ 83.32 \pm 0.02 $  & $ 47.96 \pm 0.00 $ & $ 5.99 \pm 0.00 $  \\
    SDCN  & $ 64.65 \pm 6.92 $ & $ 50.60 \pm 3.71 $  & $ 48.76 \pm 9.58 $ & $ 48.51 \pm 4.68 $  & $44.97 \pm 3.16$  & $19.86 \pm 7.16$  \\
    DMoN & $ 67.14 \pm 4.74 $ & $ 58.23 \pm 5.21 $  & $ 47.72 \pm 5.82 $ & $ 59.63 \pm 4.85 $  & $ 57.06 \pm 2.27 $  & $ 10.89 \pm 1.37 $  \\
    AGC   & $ 77.41 \pm 0.00 $  & $ 62.11 \pm 0.02 $  & $ 72.43  \pm 0.02 $ & $ 62.09 \pm 0.00 $  & $ 45.31 \pm 0.00 $  & $ 5.80 \pm 0.00 $  \\
    SSGC & $ 55.70 \pm 2.26 $ & $ 57.71 \pm 1.31 $  & $ 44.91 \pm 1.58 $ & $ 55.26 \pm 2.30 $  & $ 60.70 \pm 0.39 $  & $ 13.47 \pm 0.07 $  \\
    RwSL & $ 76.94 \pm 0.39 $ & $ 55.60 \pm 0.46 $  & $ 64.67 \pm 0.73 $ & $ \textbf{64.35} \pm 0.48 $  & 
     $ 46.26 \pm 0.72 $ &  $ 12.93 \pm 1.34 $  \\
    RwSL-R & $ {76.93}  \pm 1.13 $ & $ \textbf{62.46} \pm 1.22 $  & $  68.08  \pm 2.52 $ & $  63.30  \pm 0.95 $  & 
     $ 46.24 \pm 0.49 $ &  $ {9.00}  \pm 0.63 $  \\
     \hline
  %   \textbf{Pubmed} \\
  %  SDCN  & $ 59.95 \pm 1.00 $ & $ 17.78 \pm 0.91 $  & $ 16.39 \pm 1.16 $ & $ 60.29 \pm 1.02 $  & $55.53 \pm 0.86 $  & $ 7.50 \pm 0.58$  \\
  %  DMoN & $ 61.41 \pm 1.55 $ & $ 20.79 \pm 2.85 $  & $ 19.02 \pm 2.08 $ & $ 61.69 \pm 1.73 $  & $ 54.25 \pm 1.79 $  & $ 10.51 \pm 1.45 $  \\
  %  AGC   & $ 61.54 \pm 0.00 $  & $ 29.11 \pm 0.00 $  & $ 26.16 \pm 0.00 $ & $ 60.28 \pm 0.00 $  & $ 50.40 \pm 0.00 $  & $ 8.65 \pm 0.00 $  \\
  %  SSGC & $ \textbf{70.71} \pm 0.00 $ & $ \textbf{32.12} \pm 0.00 $  & $ \textbf{33.26} \pm 0.00 $ & $ \textbf{69.91} \pm 0.00 $  & $ \textbf{57.73} \pm 1.35 $  & $ \textbf{3.93} \pm 0.00 $  \\
  %  RwSL & $ 63.60 \pm 0.53 $ & $ 22.11 \pm 0.64 $  & $ 21.54 \pm 0.82 $ & $ 63.72 \pm 0.51 $  & 
  %   $ 57.66 \pm 0.12 $ &  $ 6.80 \pm 0.21 $  \\
  %   RwSL-R & $ 62.75 \pm 1.46 $ & $ 20.87 \pm 1.58 $  & $ 20.50 \pm 2.01 $ & $ 62.71 \pm 1.50 $  & 
  %   $ 55.08 \pm 0.81 $ &  $ 7.83 \pm 0.63 $  \\
  %   \hline
     \textbf{Friendster} \\
    RwSL & $ 76.49 \pm 2.51 $ & $ 44.20 \pm 3.73$  & $ 57.36 \pm 4.08  $ & $ 44.65 \pm 2.62 $  & 
     $ 6.48 \pm 0.16 $ &  $ 56.45 \pm 2.24 $  \\
 %   RwSL-R & $ 46.68 \pm 1.63 $ & $ 8.11 \pm 0.83$  & $ 11.42 \pm 1.36  $ & $ 23.76 \pm 0.89 $  & 
  %   $ 6.45 \pm 0.32 $ &  $ 53.15 \pm 3.48 $  \\
     \hline
    \end{tabular} 
    \end{adjustbox}
    \label{tab: Results-of-clustering-metrics}
\end{table*} 

% \vspace*{-2ex} 
\subsection{Analysis of Clustering Results - Metrics and Scalability }

% \begin{figure}
%     \centering
%     \includegraphics[scale=0.50]{Figure/ms_academic_cs_Acc_vs_Time.pdf}
%     \caption{Training Process on Coauthor CS}
%     \label{fig:Coauthor CS Training}
% \end{figure}

% \begin{figure}
%     \centering
%     \includegraphics[scale=0.50]{Figure/Train_time_scalability.pdf}
%     \caption{Scalability of Different Frameworks: Training Time vs.  Graph Size }
%     \label{fig:Time_VS_Size}
% \end{figure}

\begin{figure*}
     \centering
     \begin{subfigure}[b]{0.45\textwidth}
         \centering
         \includegraphics[width=\textwidth]{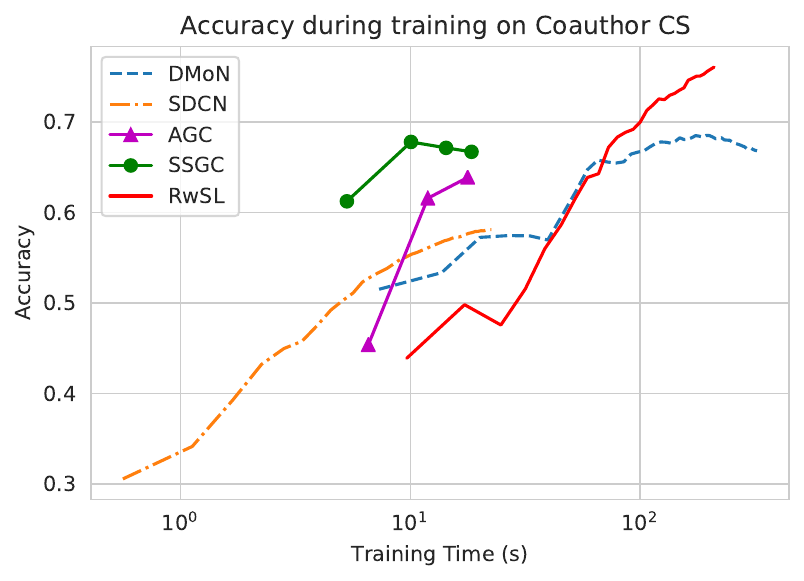}
         \caption{}
         \label{fig:Coauthor CS Training}
     \end{subfigure}
     \hfill
     \begin{subfigure}[b]{0.45\textwidth}
         \centering
         \includegraphics[width=\textwidth]{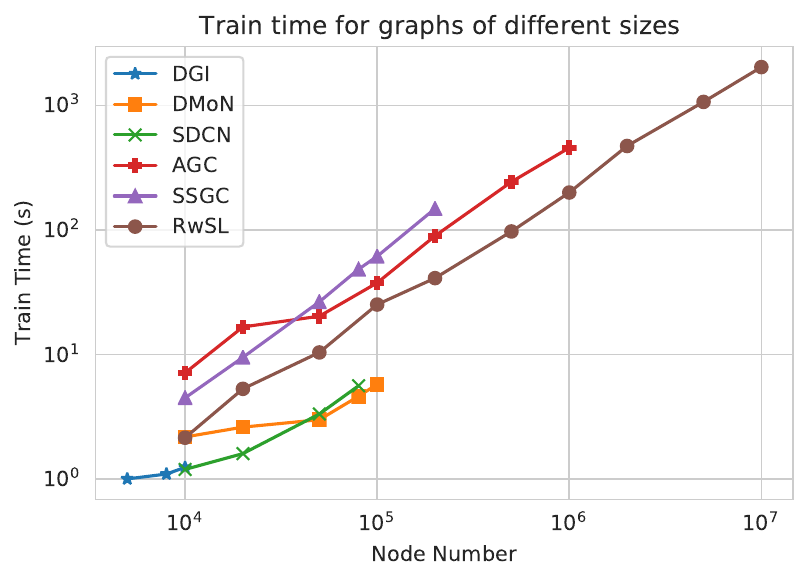}
         \caption{}
         \label{fig:Time_VS_Size}
     \end{subfigure}
     \caption{(a) Training Process on Coauthor CS; (b) Scalability of Different Frameworks: Training Time vs.  Graph Size}
\end{figure*} 

Table \ref{tab:Results-of-clustering-metrics} shows the clustering results on  seven  datasets -- each experiment is performed for 10 times with both the average and range reported. We take two settings for the input of auto-encoder yielding  two 
different versions of our method:   1) using filtered features $\Tilde{X}$ with 
resulting method referred to as {\bf  RwSL}; 2) using node features $X$ with resulting 
version referred to as {\bf  RwSL-R}.  DGI cannot handle two Coathor datasets due to high GPU memory costs.
Our RwSL based models obtained the best results for most clustering metrics for ACM, Citeseer and DBLP 
while achieving competitive results for most of the rest. RwSL avoids the over-smoothing side-effects of GCN and shows clear  advantage in clustering performance over the GCN module based methods such as SDCN \cite{Bo_2020_structural_deep} and DMoN \cite{tsitsulin2020graph_dmon}.  Figure \ref{fig:Coauthor CS Training} shows the training process of RwSL on Coauthor CS. Compared with baseline works with customized graph filters applied such as AGC~\cite{ijcai2019_AGC} and SSGC~\cite{ssgc_2021}, RwSL has additional overhead of frequent data transfers in mini-batches between devices but can achieve higher level of accuracy.

% An interesting observation  is that RwSL can out-perform RwSL-R on   Citeseer but on the contrary for ACM, DBLP, raw features lead to better result. This indicates that including  the structural information inside the autoencoder may only  provide benefits  on certain datasets. 

% Although SDCN integrates an autoencoder with a GCN module to capture structural information, it suffers from large variance and convergence difficulty as reflected by its conductance metrics on Coauthor CS. In comparison, RwSL achieved a significant clustering gain and more robustness of convergence on several datasets -- specifically, 
% for Coauthor CS, there is a  boost of 26\% on Accuracy, 25\% on NMI and 26\% on modularity; for Coauthor PHY dataset, RwSL has a metric variance around 5-12 times smaller than that of SDCN. 
 
More importantly, none of these baselines can handle large graphs. Figure~\ref{fig:Time_VS_Size}  illustrates the scalability landscape of the state-of-the-art methods. Specifically,  we generated synthetic  undirected graphs with an increasing number of nodes  using PaRMAT \cite{PaPMAT_2015} with edge size set as 20 times the number of nodes and a random feature matrix with a dimension of 1000.  We performed 5-epoch training for each framework.
and observed all baselines can only handle graphs with a maximum  of hundreds  of thousands of nodes, as limited by the GPU memory capacity. In comparison, 
our proposed method continues  to scale linearly. We further performed the node clustering on a billion-scale graph Friendster following the settings as  used by authors of  GnnBP \cite{cwdlydw2020gbp}. Similar settings have been adopted in several prior works on community detection \cite{Net_community_Leskovec_2010,Heat_kernel_Kloster_2014,pagerank_heat_kernel_yang_2019}. 

% Note that no comparison is feasible against other frameworks as none of them could process this graph on a GPU. 

\subsection{Node Embedding t-SNE}

\begin{figure*}
     \centering
     \begin{subfigure}[b]{0.32\textwidth}
         \centering
         \includegraphics[width=\textwidth]{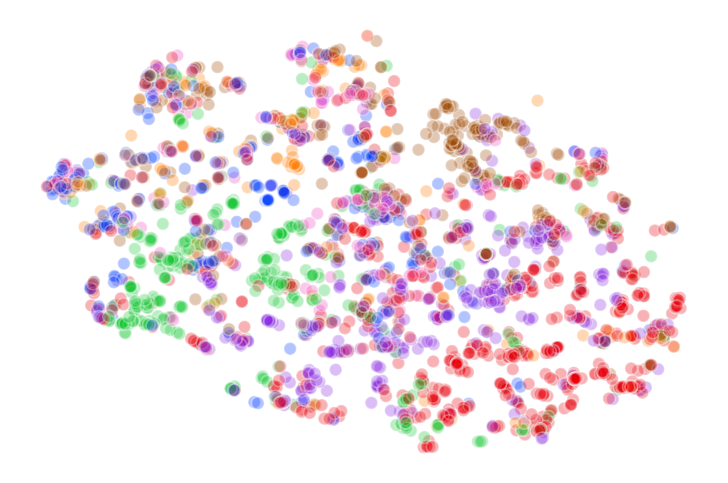}
         \caption{Raw node attributes}
     \end{subfigure}
     \hfill
     \begin{subfigure}[b]{0.32\textwidth}
         \centering
         \includegraphics[width=\textwidth]{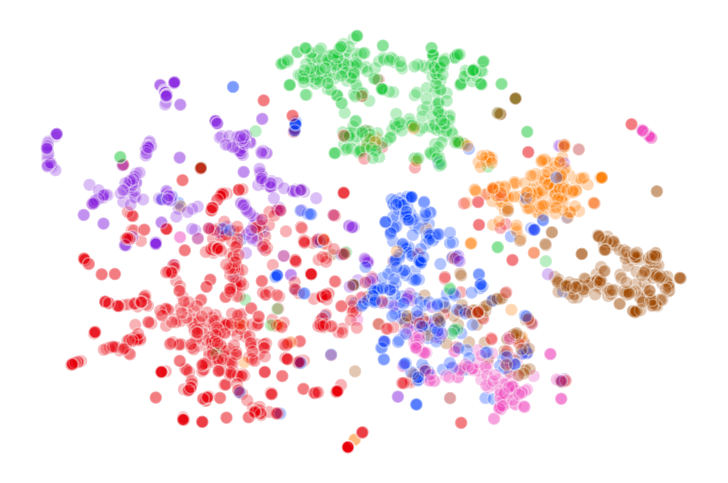}
         \caption{PPR filtered attributes}
     \end{subfigure}
     \hfill
     \begin{subfigure}[b]{0.32\textwidth}
         \centering
         \includegraphics[width=\textwidth]{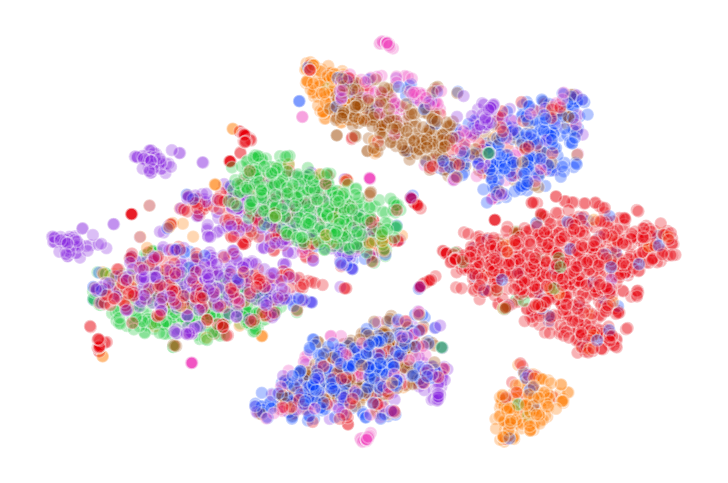}
         \caption{Encoder}
     \end{subfigure}
    \caption{Cora t-SNE for different features/embeddings. Each color represents a distinct class.}
    \label{fig: cora tsne embedding model}
\end{figure*}

Figure \ref{fig: cora tsne embedding model} intuitively shows comparison between node embeddings with node attributes using t-SNE algorithm \cite{t-SNE_algo}. We can see that the graph filter and encoder module can gradually ease the clustering tasks by distinguishing clusters with less overlapping areas.

\subsection{Training Convergence}

\begin{figure}
    \centering
    \includegraphics[width=\linewidth]{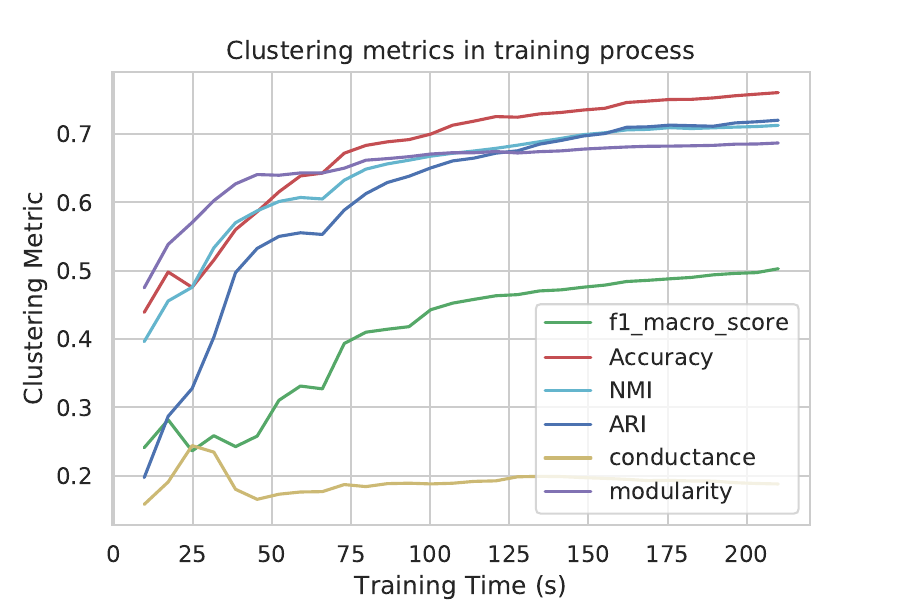}
     \caption{Clustering Metrics Convergence Curve}
     \label{fig: Coauthor_CS Clustering metrics RwSL}
\end{figure}

 RwSL achieves robust convergence across all real-word datasets as used in this work. For example, Figure \ref{fig: Coauthor_CS Clustering metrics RwSL} shows the training process of RwSL on Coauthor CS with all metrics converged to a steady state.

\section{Details for Experimental Setup}

\subsection{Experimental Settings}
Our RwSL framework is implemented in PyTorch 1.7 on CUDA 10.1, whereas   the graph filtering procedure is in C++. Our experiments are performed on nodes with a dual Intel Xeon 6148s @2.4GHz CPU and dual NVIDIA Volta V100 w/ 16GB memory GPU and 384 GB DDR4 memory. 
Graph filtering is executed on CPU while both autoencoder pretraining and DNN co-train process are performed on a single GPU. We used dropout to make our model more robust and applied a AdamW optimization method with a decoupled weight decay regularization technique \cite{loshchilov2019decoupled_weight_decay}. 

\subsection{RwSL Hyper-parameter Settings}\label{RwSL hyper-parameter}

\begin{table*}
    \centering
    \caption{ RwSL hyper-parameter settings on 6 datasets}
    \begin{adjustbox}{width=\textwidth}
    \begin{tabular}{lrrrrrr}
    \hline
    Hyper-parameters  & ACM & CiteSeer & DBLP & Cora & Coauthor CS & Coauthor PHY \\
    \hline
    learning\_rate     & $ 1e-4 $   & $ 1e-4 $   & $ 1e-4 $  & $ 1e-4 $ & $ 1e-5 $ & $ 1e-4 $  \\
    architecture & 512-512-2048-16  & 512-512-2048-16 & 512-8 & 512-2048-32 & 512-2048-16 & 512-1024-8 \\
    n\_epochs          & $ 100 $    & $  120 $   & $ 100 $  & $ 100 $ & $ 30 $ & $  20 $  \\
    $v$                 &  $ 1.0$   &  $ 1.0$    &  $ 1.0$  &  $ 1.0$ & $ 1.0$ &  $ 1.0$ \\
    $\beta$           & $0.01$      & $0.01$    & $0.01$    & $0.01$ & $0.05$  & $0.05$ \\
    $\gamma$           & $0.1$      & $0.1$    & $0.1$      & $0.1$ & $0.2$ & $0.2$  \\
    dropout\_rate      & $ 0.01 $   & $ 0.01 $  & $ 0.01 $  & $ 0.01 $  & $ 0.01 $  & $ 0.01 $  \\
    weight\_decay      & $ 0.01 $   & $ 0.01 $  & $ 0.01 $  & $ 0.01 $  & $ 0.5 $ & $0.01$   \\
    update\_p           & $ 1 $       & $ 1 $     & $ 1 $   & $ 1 $  & $ 1 $  & $ 1 $  \\
    pretrain\_lr        & $ 1e-4 $   & $ 1e-4 $  & $ 1e-4 $ & $ 1e-4 $   & $ 1e-4 $  & $ 1e-4 $ \\
    pretrain\_n\_epochs & $ 60 $     & $ 60  $     & $ 60 $ & $ 30 $  & $ 60 $  & $ 30 $  \\
    $\alpha$       & $ 0.1 $   & $ 0.1 $   & $ 0.1 $  & $ 0.1 $  & $ 0.1 $  & $ 0.1 $  \\
    $r_{max}$                & $ 1e-5 $   & $ 1e-5 $  & $ 1e-5 $ & $ 1e-5 $   & $ 1e-5 $  & $ 1e-5 $ \\
    rrz                 & $ 0.4 $   & $ 0.4 $   & $ 0.4 $   & $ 0.4 $ & $ 0.4 $ & $ 0.4 $  \\
    \hline
    \end{tabular}
    \end{adjustbox}
    \label{tab: RwSL datasets hyper-parameters}
\end{table*}

Detailed hyper-parameters settings are included in Table \ref{tab: RwSL datasets hyper-parameters}. Learning\_rate and pretrain\_lr correspond to the learning rates during DNN training and autoencoder training. And n\_epochs and pretrain\_n\_epochs are the number of iterations for DNN training and autoencoder pretrain respectively. The architecture describes the number of neurons on each layer. We set the degree of freedom of the Student t-distribution $v$ as 1.0 globally. $\beta$ and $\gamma$ are the weighting factors for the training objective function. We applied a decoupled weight decay regularization \cite{loshchilov2019decoupled_weight_decay} resulting in the factor weight\_decay and dropout technique with the dropout rate described by dropout\_rate. The period (update\_p) to update target distribution P is fixed as 1. The last three parameters are from GnnBP framework \cite{chen2020scalable}, $\alpha \in  (0, 1)$ is teleport probability defined in
Personalized PageRank weights $(w_{l}=\alpha(1-\alpha)^{l})$; $r_{max}$ is the threshold during reverse push propagation from the feature vectors; $rrz$ is the convolutional coefficient.

%% file: text/Conc.tex
% \vspace*{-2ex} 
\section{Conclusions} 
% \vspace*{-2ex}  

This paper proposed a Random-walk based Scalable Learning (RwSL), which essentially incorporates a Laplacian smoothing based graph filter by using 
 a  random-walk based algorithm to approximate Generalized PageRank (GPR). This is made feasible by a novel derivation that establishes a  correspondence between Laplacian smoothing and Generalized PageRank (GPR). Our additional  theoretical analysis provides spectral properties of a GPR based graph filter and impact of major 
 parameters on clustering performance. Our extensive evaluation has shown that for 
 small and medium sized graphs, we 
 produce results that are  better or competitive over the existing 
 methods. Unlike all other methods, we are able to perform deep 
 clustering on a graph with  1.8 billion edges.